\documentclass[journal]{IEEEtran}

\usepackage{amsmath,engord}
\usepackage{amsfonts}
\usepackage{subfigure}
\usepackage{algorithm,algorithmic}
\usepackage{setspace,bbm,color}
\usepackage{cite,url,subfigure,epsfig,graphicx,makecell,multirow}
\usepackage{times,amssymb,verbatim,amsfonts,amsmath,mathrsfs,threeparttable,booktabs}

\makeatletter
\def\ps@headings{%
\def\@oddfoot{}%
\def\@evenfoot{}}
\makeatother
\newcommand{\squishlist}{
   \begin{list}{$\bullet$}
    { \setlength{\itemsep}{0pt}      \setlength{\parsep}{3pt}
      \setlength{\topsep}{3pt}       \setlength{\partopsep}{0pt}
      \setlength{\leftmargin}{1.5em} \setlength{\labelwidth}{1em}
      \setlength{\labelsep}{0.5em} } }
\newcommand{\squishend}{\end{list}  }

\newcommand{\tabincell}[2]{\begin{tabular}{@{}#1@{}}#2\end{tabular}}%

\begin{document}

\title{Deep Reinforcement Learning for Traffic Light Control in Vehicular Networks}

\author{Xiaoyuan~Liang,
	Xusheng Du, ~\IEEEmembership{Student Member,~IEEE,}
        Guiling Wang, ~\IEEEmembership{Member,~IEEE,}
        and~Zhu~Han ~\IEEEmembership{Fellow,~IEEE}
\thanks{X. Liang and G. Wang are with the Department
of Computer Science, New Jersey Institute of Technology, Newark,
NJ, 07102 USA email: \{xl367, gwang\}@njit.edu.}
\thanks{X. Du and Z. Han is with Department of Electrical and Computer Engineering, University of Houston, Houston, TX 77004 USA email: \{xunshengdu, hanzhu22\}@gmail.com.}
\thanks{Manuscript received January 22, 2018.}}

\markboth{IEEE Transactions on Vehicular Technology,~Vol.~XX, No.~XX}%
{Liang \MakeLowercase{\textit{et al.}}: Deep Reinforcement Learning for Traffic Light Control in Vehicular Networks}

\maketitle

\begin{abstract}
Existing inefficient traffic light control causes numerous problems, such as long delay and waste of energy. 
To improve efficiency, taking real-time traffic information as an input and dynamically adjusting the 
traffic light duration accordingly is a must. 
In terms of how to dynamically adjust traffic signals' duration, 
existing works either split the traffic signal into equal duration 
or extract limited traffic information from the real data. 
In this paper, we study how to decide the traffic signals' duration 
based on the collected data from different sensors and vehicular networks.
We propose a deep reinforcement learning model to control the traffic light. 
In the model, we quantify the complex traffic scenario as states by collecting data and dividing the whole intersection into small grids.
The timing changes of a traffic light are the actions, which are modeled as a high-dimension Markov decision process.
The reward is the cumulative waiting time difference between two cycles. 
To solve the model, a convolutional neural network is employed to map the states to rewards.
The proposed model is composed of several components to improve the performance, such as dueling network, target network, double Q-learning network, and prioritized experience replay.
We evaluate our model via simulation in the Simulation of Urban MObility (SUMO) in a vehicular network, and
the simulation results show the efficiency of our model in controlling traffic lights.
\end{abstract}

\section{Introduction}
\label{sec:intr}

Existing road intersection management is done through traffic lights. 
The inefficient traffic light control causes numerous problems,
such as long delay of travelers, huge waste of energy and worsening air quality. 
In some cases, it may also contribute to vehicular accidents \cite{MouSch17,GenRaz16}.
Existing traffic light control either deploys fixed programs without considering real-time traffic 
or considering the traffic to a very limited degree\cite{Cas17}.
The fixed programs set the traffic signals equal time duration in every cycle, 
or different time duration based on historical information.
Some other control programs take inputs from sensors such as underground inductive loop detectors
to detect the existence of vehicles in front of traffic lights. 
The inputs are processed in a very coarse way to determine the duration of green/red lights. 

In some cases, existing traffic light control systems work, though at a low efficiency. 
However, in many other cases, such as a football event or a more common high traffic hour scenario,
the traffic light control systems become paralyzed.
Instead, we often witness policemen directly manage the intersection by waving signals. 
This human operator can see the real time traffic condition in the intersecting roads and smartly 
determine the duration of the allowed passing time for each direction using his/her long-term experience and understanding
about the intersection. 
The operation normally is very effective. 
The witness motivates us to propose a smart intersection traffic light management system which 
can take real-time traffic condition as input and learn how to manage the intersection just like the human operator. 

To implement such a system, we need `eyes' to watch the real-time road condition and `a brain' to process it. 
For the former, recent advances in sensor and networking technology enables taking real-time traffic information 
as input, such as the number of vehicles, the locations of vehicles, and their waiting time\cite{ElAbd14}. 
For the `brain' part, reinforcement learning, as a type of machine learning techniques, is a promising 
way to solve the problem.
A reinforcement learning system's goal is to make an action agent learn the optimal policy 
in interacting with the environment to maximize the reward, e.g., 
the minimum waiting time in our intersection control scenario. 
It usually contains three components, states of the environment, action space of the agent, 
and reward from every action\cite{SutBar98}.
A well-known application of reinforcement learning is AlphaGo \cite{SliHua16}, including AlphaGo Zero \cite{SilSch17}.
AlphaGo, acting as the action agent in a Go game (environment), first observes the current image of the chessboard (state), and takes the image as the input of a reinforcement learning model to determine where to place the optimal next playing piece `stone' (action).
Its final reward is to win the game or to lose.
Thus, the reward may be unobvious during the playing process and it is delayed till the game is over.
When applying reinforcement learning to the traffic light control problem, 
the key point is to define the three components at an intersection and quantify them to be computable. 

Some researchers have proposed to dynamically control the traffic lights using reinforcement learning. 
Early works define the states by the number of waiting vehicles or the waiting queue length \cite{ElAbd14, AbdMoz13}. 
But real traffic situation cannot be accurately captured by the number of waiting vehicles or queue length \cite{GenRaz16}.
With the popularization of vehicular networks and cameras, more information about roads can be extracted and transmitted via the network, such as vehicles' speed and waiting time \cite{HarLab08}. 
However, more information causes the dramatically increasing number of states. 
When the number of states increases, the complexity in a traditional reinforcement learning system grows exponentially.
With the rapid development of deep learning, deep neural networks have been employed to deal with the large number of states, which 
constitutes a deep reinforcement learning model \cite{MniKav15}.
A few recent studies have proposed to apply deep reinforcement learning in the traffic light control problem \cite{LiLv16,Van16}.
But there are two main limitations in the existing studies: (1) the traffic signals are usually split into fixed-time intervals, and the duration of green/red lights can only be a multiple of this fixed-length interval, 
which is not efficient in many situations;  
(2) the traffic signals are designed to change in a random sequence, 
which is not a safe nor comfortable way for drivers.
In this paper, we study the problem on how to control the traffic light's signal duration in a cycle based on the extracted information from vehicular networks to help efficiently manage vehicles at an intersection.

In this paper, we solve the problem in the following approaches and make the following contributions.
Our general idea is to mimic an experienced operator to control the signal duration in every cycle based on the information gathered from vehicular networks.
To implement such an idea, the experienced operator's operation is modeled as an Markov Decision Process (MDP). 
The MDP is a high-dimension model, which contains the time duration of every phase. 
The system then learns the control strategy based on the MDP by trial and error in a deep reinforcement learning model. 
To fit a deep reinforcement learning model, we divide the whole intersection into grids and build a matrix from the vehicles' information in the grids collected by vehicular networks or extracted from a camera via image processing.
The matrix is defined as the states and the reward is the cumulative waiting time difference between two cycles.
In our model, a convolutional neural network is employed to match the states and expected future rewards.
In the traffic light control problem, every traffic light's action may affect the environment and the traffic flow changes dynamically, which makes the environment unpredictable.
Thus, a convolutional network is hard to predict the accurate reward.
Inspired by the recent studies in reinforcement learning, we employ a series of state-of-the-art techniques in our model to improve the performance, including dueling network \cite{WanSch15}, target network \cite{MniKav15}, double Q-learning network \cite{VanGue16}, and prioritized experience replay \cite{SchQua15}.
In this paper, we combine these techniques as a framework to solve our problem, which can be easily applied into other problems. 
Our system is tested on a traffic micro-simulator, Simulation of Urban MObility (SUMO) \cite{KraErd12}, and the simulation results show the effectiveness and high-efficiency of our model.

The reminder of this paper is organized as follows.
The literature review is presented in Section \ref{sec:lit}.
The model and problem statement are introduced in Section \ref{sec:model}.
 The background on reinforcement learning is introduced in Section \ref{sec:brl}.
Section \ref{sec:rlm} shows the details in modeling an reinforcement learning model in the traffic light control system of vehicular networks.
Section \ref{sec:dqn} extends the reinforcement learning model into a deep learning model to handle the complex states in the our system.
The model is evaluated in Section \ref{sec:eva}.
Finally, the paper is concluded in Section \ref{sec:con}.

\section{Literature Review}
\label{sec:lit}
Previous works have been done to dynamically control adaptive traffic lights.
But due to the limited computing power and simulation tools, early studies focus on solving the problem by fuzzy logic\cite{ChiCha93}, linear programming \cite{De99},
etc.\
In these works, road traffic is modeled by limited information,
which cannot be applied in large scale.

Reinforcement learning was applied in traffic light control since 
1990s.
El-Tantawy \textit{et al.} \cite{ElAbd14} summarize the methods from 1997 to 2010 that use reinforcement learning to control traffic light timing.
During this period, the reinforcement learning techniques are limited to tabular Q learning and a linear function is normally used to estimate the Q value. 
Due to the technique limitation at the time in reinforcement learning, they usually make a small-size state space, such as the number of waiting vehicles \cite{AbdPri03, ChiBol11, AbdMoz13} and the statistics of traffic flow \cite{AreLiu10, BalGer10}.
The complexity in a traffic road system can not be actually presented by such limited information.
When much useful relevant information is omitted in the limited states, it seems unable to act optimally in traffic light control\cite{GenRaz16}.


With the development of deep learning and reinforcement learning, they are combined together as deep reinforcement learning to estimate the Q value. 
We summarize the recent studies that use the value-based deep reinforcement learning to control traffic lights in Table \ref{tbl:rvw}.
There are three limitations in these previous studies.
Firstly, most of them test their models in a simple cross-shape intersection with through traffic only \cite{LiLv16, Van16}.
Secondly, none of the previous works determines the traffic signal timing in a whole cycle.
Thirdly, deep reinforcement learning is a fast developing field, where a lot of new ideas are proposed in these two years, such as dueling deep Q network \cite{WanSch15}, but they have not been applied in traffic control.
In this paper, we make the following progress.
Firstly, our intersection scenario contains multiple phases, which corresponds a high-dimension action space in a cycle.
Secondly, our model guarantees that the traffic signal time smoothly changes between two neighboring actions, which is exactly defined in the MDP model.
Thirdly, we employ the state-of-the-art techniques in value-based reinforcement learning algorithms to achieve good performance, which is evaluated via simulation.

\begin{table*}
\centering
\caption{\textbf{List of previous studies that use value-based deep reinforcement learning to adaptively control traffic signals}}
\label{tbl:rvw}
\begin{tabular}{lcccccc}
    \Xhline{1\arrayrulewidth}
 Study & State & Action & Reward &  Time step & Note \\
\hline
\tabincell{l}{Genders \textit{et al.} \\ (2016) \cite{GenRaz16}}& \tabincell{c}{Position\\ speed} & 4 phases & \tabincell{c}{Change in \\ cumulative delay} & NA & \tabincell{c}{Convolutional neural network}\\

    \tabincell{l}{Li \textit{et al.} \\ (2016) \cite{LiLv16}}& Queue length & 2 phases & \tabincell{c}{Difference between \\ flows in two directions} &  5s & \tabincell{c}{Stacked auto-encoders}\\

\tabincell{l}{Van Der Pol \\ (2016) \cite{Van16}}& Position & 2 phases & \tabincell{c}{ Teleport, wait time,\\ stop, switch, and delay} & 1s & \tabincell{c}{Double Q network \\ Prioritized experience replay}\\



\tabincell{l}{Gao \textit{et al.} \\ (2017) \cite{GaoShe17}}& \tabincell{c}{Position \\ speed} & 4 phases & \tabincell{c}{Change in \\ cumulative staying time} & 6/10s & \tabincell{c}{Convolutional neural network\\ Experience replay}\\
    \Xhline{1\arrayrulewidth}
\end{tabular}
\end{table*}

\section{Model and Problem Statement}
\label{sec:model}

\begin{figure}
\centerline{ \psfig{figure=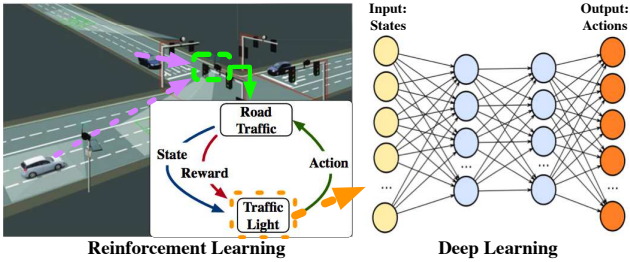,width=3.3in}}
    \caption{The traffic light control model in our system. The left side shows the intersection scenario where the traffic light gathers vehicles' information via a vehicular network and it is controlled by the reinforcement learning model; the right side shows a deep neural network to help the traffic light choose an action.}
\label{fig:mdl}
\end{figure}

In this paper, we consider a road intersection scenario where traffic lights are used to control traffic flows.
The model is shown in Fig. \ref{fig:mdl}.
The left side shows the structure in a traffic light. 
The traffic light first gathers road traffic information via a vehicular network \cite{HarLab08}, which is presented by the dashed purple lines in the figure.
The traffic light processes the data to obtain the road traffic's state and reward, which has been assumed in many previous studies \cite{GenRaz16, Van16, GaoShe17}.
The traffic light chooses an action based on the current state and reward using a deep neural network shown in the right side.
The left side is the reinforcement learning part and the deep learning part.
They make up our deep reinforcement learning model in traffic light control.

In our model, traffic lights are used to manage the traffic flows at intersections.
A traffic light at an intersection has three signals, green, yellow and red.
One traffic light may not be enough to manage all the vehicles when there are vehicles from multiple directions at an intersection. 
Thus, multiple traffic lights need to cooperate at a multi-direction intersection.
At such an intersection, the traffic signal guides vehicles from non-conflicting directions at one time by changing the traffic lights' statuses. 
One status is one of all the legal combinations of all traffic lights' red and green signals omitting the yellow signals.
The time duration staying at one status is called one phase.
The number of phases is decided by the number of legal statuses at an intersection.
All the phases cyclically change in a fixed sequence to guide vehicles to pass the intersection.
It is called a cycle when the phases repeat once.
The sequence of phases in a cycle is fixed, but the duration of every phase is adaptive.
If one phase needs to be skipped, its duration can be set 0 second.
In our problem, we dynamically adjust the duration in every phase to deal with different traffic situations at an intersection. 

Our problem is defined by how to optimize the efficiency of the intersection usage by dynamically changing every phase's duration of a traffic light via learning from historical experiences.
The general idea is to extend the duration for the phase that has more vehicles in that direction.
But it is time-consuming to train a person to become a master who well knows how much time should be given to a phase based on current traffic situation.
Reinforcement learning is a possible way to learn how to control the traffic light and liberate a human being from the learning process.
Reinforcement learning updates its model by continuously receiving states and rewards from the environment.
The model gradually becomes a mature and advanced model.
It is different from supervised learning in not requiring numerous data at one time.
In this paper, we employ the deep reinforcement learning to learn the timing strategy of every phase to optimize the traffic management. 

\section{Background on Reinforcement Learning}
\label{sec:brl}
Reinforcement learning is one category of algorithms in machine learning, which is different from supervised learning and unsupervised learning \cite{SutBar98}.
It interacts with the environment to get rewards from actions.
Its goal is to take the action to maximize the numerical rewards in the long run.
In reinforcement learning, an agent, the action executor, takes an action and the environment returns a numerical reward based on the action and current state.
A four-tuple $\left \langle S, A, R, T \right \rangle$ can be used to denote the reinforcement learning model with the following meanings:
\begin{itemize}
\item $S:$ the possible state space. $s$ is a specific state ($s \in S$);
\item $A:$ the possible action space. $a$ is an action ($a \in A$);
\item $R:$ the reward space. $r_{s,a}$ means the reward in taking action $a$ at state $s$;
\item $T:$ the transmission function space among all states, which means the probability of the transmission from one state to another. 
\end{itemize}
In a deterministic model, $T$ is usually omitted. 

A policy is made up of a series of consequent actions. 
The goal in reinforcement learning is to learn an optimal policy to maximize the cumulative expected rewards starting from the current state.
Generally speaking, the agent at one specific state $s$ takes an action $a$ to reach state $s'$ and gets a reward $r$, which is denoted by $\langle s, a, r, s' \rangle$.
Let $t$ denote the $t$\textsuperscript{th} step in the policy $\pi$.
The cumulative reward in the future by taking action $a$ at state $s$ is defined by $Q(s,a)$ in the following equation, 
\begin{equation}
\begin{aligned}
    Q^{\pi}(s, a) &= E\left [r_t+\gamma r_{t+1}+ \gamma^2 r_{t+2}+ \cdots | s_t=s, a_t=a, \pi\right]&\\
    &=E\left[\sum_{k=0}^{\infty} \gamma^k r_{t+k} | s_t = s, a_t=a, \pi \right].&
\end{aligned}
\end{equation}
In the equation, 
$\gamma$ is the discount factor, which is usually in $\left[0, 1\right)$.
It means the nearest rewards are worthier than the rewards in the further future.

The optimal action policy $\pi^*$ can be obtained recursively.
If the agent knows the optimal $Q$ values of the succeeding states, the optimal policy just chooses the action that achieves the highest cumulative reward.
Thus, the optimal $Q(s, a)$ is calculated based on the optimal $Q$ values of the succeeding states.
It can be expressed by the Bellman optimality equation to calculate $Q^{\pi^*}(s, a)$,
\begin{equation}
\label{equ:blm}
Q^{\pi^*}(s, a) = E_{s'}\left[r_{t}+\gamma \max_{a'}Q^{\pi^*}(s', a') |s, a\right].
\end{equation}
The intuition is that the cumulative reward is equal to the sum of the immediate reward and optimal future reward thereafter.
If the estimated optimal future reward is obtained, the cumulative reward since now can be calculated.
This equation can be solved by dynamic programming, but it requires that the number of states is finite to make the computing complexity controllable. 
When the number of states becomes large, a function $\theta$ is needed to approximate the $Q$ value, which will be shown in Section \ref{sec:dqn}.



\section{Reinforcement Learning Model}
\label{sec:rlm}
To build a traffic light control system using reinforcement learning, we need to define the states, actions and rewards. 
In the reminder of this section, we present how the three elements are defined in our model. 

\subsection{States}
We define the states based on two pieces of information, 
position and speed of vehicles at an intersection.
Through a vehicular network, vehicles' position and speed can be obtained \cite{HarLab08}.
Then the traffic light can extract a virtual snapshot image of the current intersection. 
The whole intersection is divided into same-size small square-shape grids.
The length of grids, $c$, should guarantee that no two vehicles can be held in the same grid and one entire vehicle can be put into a grid to reduce computation.
The value of $c$ in our system will be given in the evaluation.
In every grid, the state value is a two-value vector $<position, speed>$ of the inside vehicle.
The position dimension is a binary value, which denotes whether there is a vehicle in the grid.
If there is a vehicle in a grid, the value in the grid is 1; otherwise, it is 0.
The speed dimension is an integer value, denoting the vehicle's current speed in $m/s$.

\begin{figure}[h!]
\hspace*{\fill}%
\centerline{\subfigure[The snapshot of traffic on a road at one moment]{\label{fig:st1}\includegraphics[width=0.45\textwidth]{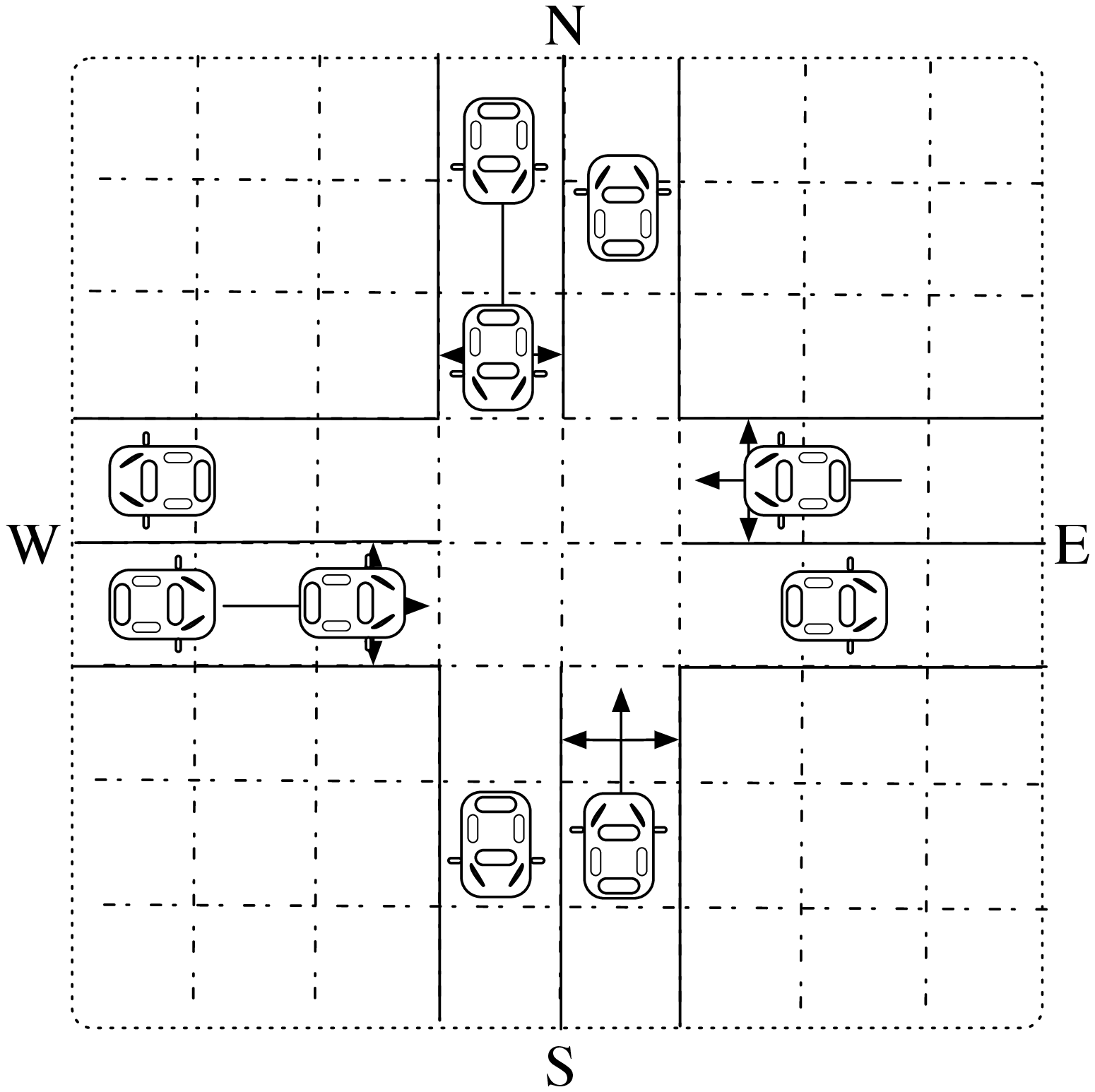}}}\\
\centerline{\subfigure[The corresponding position matrix on this road]{\label{fig:st2}\includegraphics[width=0.4\textwidth]{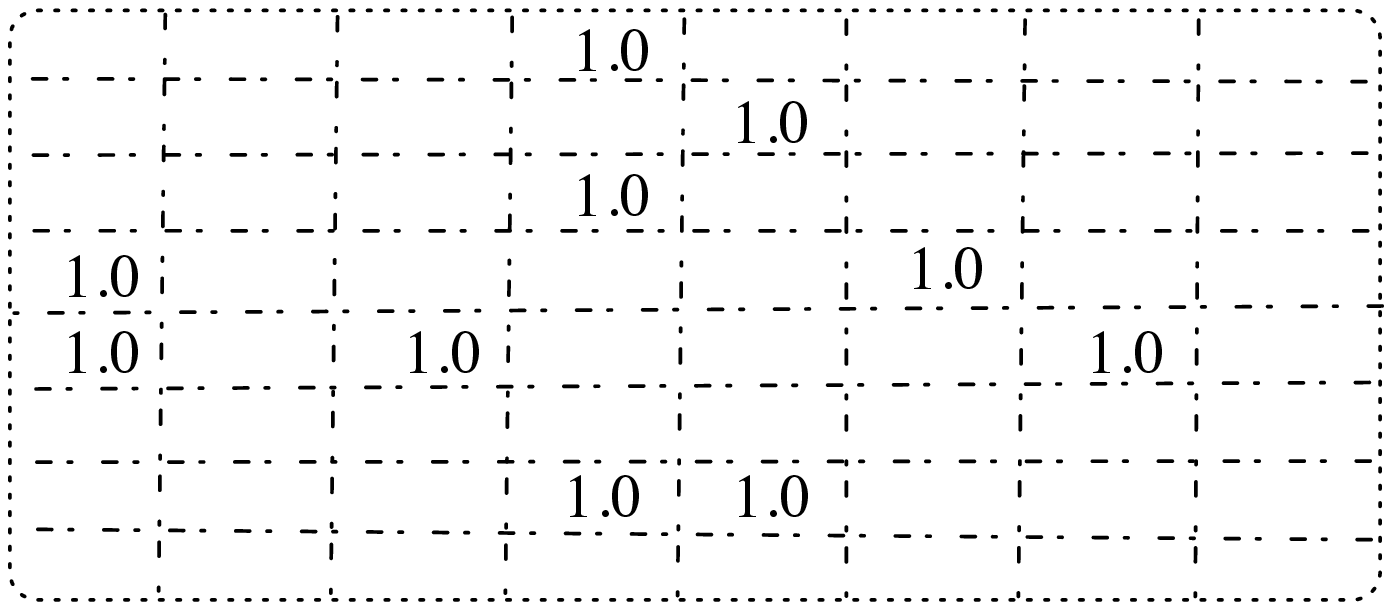}}}
\hspace*{\fill}%
\caption{The process to build the state matrix.}
\label{fig:st}
\end{figure}

Let's take Fig. \ref{fig:st} as an example to show how to quantify the intersection to obtain the state values.
Fig. \ref{fig:st1} shows a snapshot of the traffic status at a simple one-lane four-way intersection, which is built with information in a vehicular network.
The intersection is split into square-shape grids.
The position matrix has the same size of the grids, which is shown in Fig. \ref{fig:st2}.
In the matrix, one cell corresponds to one grid in Fig. \ref{fig:st1}.
The blank cells mean no vehicle in the corresponding grid, which are $0$. 
The other cells with vehicles inside are set $1.0$.
The value in the speed dimension is built in a similar way.
If there is a vehicle in the grid, the corresponding value is the vehicle's speed; otherwise, it is 0.

\subsection{Action Space}
A traffic light needs to choose an appropriate action to well guide vehicles at the intersection based on the current traffic state.
In this system, the action space is defined by selecting every phase's duration in the next cycle.
But if the duration changes a lot between two cycles, the system may become unstable. 
Thus, the legal phases' duration at the current state should smoothly change. 
We model the duration changes of legal phases between two neighboring cycles as a high-dimension MDP.
In the model, the traffic light only changes one phase's duration in a small step.

\begin{figure}
\centerline{ \psfig{figure=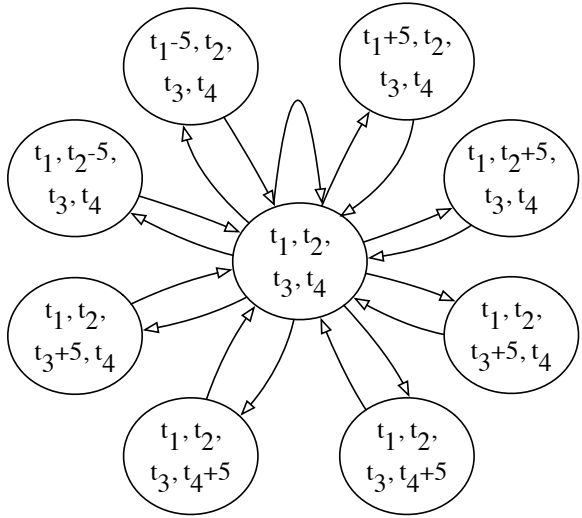,width=3.0in}}
\caption{Part of the Markov decision process in a multiple traffic lights scenario.}
\label{fig:mdp2}
\end{figure}

Let's take the intersection in Fig. \ref{fig:st1} as an example.
At the intersection, there are four phases, north-south green, east-north\&west-south green, east-west green, and east-south\&west-north green.
The other unmentioned directions are red by default.
Let's omit the yellow signals here, which will be presented later.
Let a four-tuple $<t_1, t_2, t_3, t_4>$ denote the duration of the four phases in current cycle. 
The legal actions in the next cycle is shown in Fig. \ref{fig:mdp2}.
In the figure, one circle means the durations of the four phases in one cycle.
We discretize the time change from the current cycle to the succeeding cycle to 5 seconds.
The duration of one and only one phase in the next cycle is the current duration added or subtracted by 5 seconds.
After choosing the phases' duration in the next cycle, the current duration becomes the chosen one.
The traffic light can select an action in a similar way as the previous procedure.
In addition, we set 
the max legal duration of a phase as 60 seconds and the minimal as 0 second.

The MDP is a flexible model.
It can be applied into a more complex intersection with more traffic lights, which needs more phases, such as an irregular intersection with five or six ways.
When there are more phases at an intersection, they can be added in the MDP model as a higher-dimension value.
The dimension of the circle in the MDP is equal to the number of phases at the intersection.

The phases in a traffic light cyclically change in a sequence. 
Yellow signal is required between two neighboring phases to guarantee safety, which allows running vehicles to stop before signals become red.
The yellow signal duration $T_{yellow}$ is defined by the maximum speed $v_{max}$ on that road divided by the most commonly-seen decelerating acceleration $a_{dec}$.
\begin{equation}
T_{yellow} = \frac{v_{max}}{a_{dec}}.
\end{equation}
It means the running vehicle needs such a length of time to firmly stop in front of the intersection.


\subsection{Rewards}
Rewards are an element that differentiates reinforcement learning from other learning algorithms.
The role of rewards is to provide feedback to a reinforcement learning model about the performance of the previous actions.
Thus, it is important to define the reward to correctly guide the learning process, which accordingly helps take the best action policy.

In our system, the main goal is to increase the efficiency of an intersection.
A main metric in the efficiency is vehicles' waiting time.
Thus, we define the rewards as the change of the cumulative waiting time between two neighboring cycles.
Let $i_t$ denote the $i$\textsuperscript{th} observed vehicle from the starting time to the starting time point of the  $t$\textsuperscript{th} cycle and $N_t$ denote the corresponding total number of vehicles till the $t$\textsuperscript{th} cycle.
The waiting time of vehicle $i$ till the $t$\textsuperscript{th} cycle is denoted by $w_{i_t,t}, (1\leq i_t\leq N_t)$.
The reward in the $t$\textsuperscript{th} cycle is defined by the following equation,  
\begin{equation}
r_t = W_t-W_{t+1}, 
\end{equation}
where
\begin{equation}
W_t = \sum_{i_t=1}^{N_t} w_{i_t,t}.
\end{equation}
It means the reward is equal to the increment in cumulative waiting time between before taking the action and after the action.
If the reward becomes larger than before, the waiting time increases less than before. 
Considering the delay is non-decreasing with time, the overall reward is always non-positive.

\section{Double Dueling Deep Q Network}
\label{sec:dqn}
There are a lot of practical problems in directly solving (\ref{equ:blm}), such as the states are required to be finite \cite{SutBar98}.
In the traffic light control system in vehicular networks, the number of states are too large.
Thus, in this paper we propose a Convolutional Neural Network (CNN) \cite{LiaWan17} to approximate the $Q$ value.
Combining with the state-of-the-art techniques, the proposed whole network is called Double Dueling Deep Q Network (3DQN).

\subsection{Convolutional Neural Network}
\begin{figure}
\centerline{ \psfig{figure=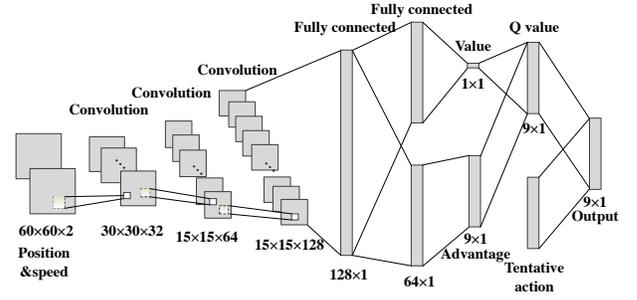,width=3.2in}}
\caption{The architecture of the deep convolutional neural network to approximate the $Q$ value.}
\label{fig:cnn}
\end{figure}
The architecture of the proposed CNN is shown in Fig. \ref{fig:cnn}.
It is composed of three convolutional layers and several fully-connected layers.
In our system, the input is the small grids including the vehicles' position and speed information. 
The number of grids at an intersection is $60\times 60$. 
The input data become $60\times60\times2$ with both position and speed information.
The data are first put through three convolutional layers. 
Each convolutional layer includes three parts, convolution, pooling and activation.
The convolutional layer includes multiple filters.
Every filter contains a set of weights, which aggregates local patches in the previous layer and shifts a fixed length of step defined by the stride each time.
Different filters have different weights to generate different features in the next layer.
The convolutional operation makes the presence of a pattern more important than the pattern's position.
The pooling layer selects the salient values from a local patch of units to replace the whole patch.
The pooling process removes less important information and reduces the dimensionality.
The activation function is to decide how a unit is activated. 
The most common way is to apply a non-linear function on the output.
In this paper, we employ the leaky ReLU \cite{HeZha15} as the activation function with the following form (let $x$ denote the output from a unit), 
\begin{equation}
    \label{equ:act}
    f(x) = \begin{cases} x, & $if $ x>0, \\
            \beta x, & $if $ x\leq 0.\\
    \end{cases}
\end{equation}
$\beta$ is a small constant to avoid zero gradient in the negative side.
The leaky ReLU can converge faster than other activation functions, like tanh and sigmoid, and prevent the generation of `dead' neurons from regular ReLU.

In the architecture, three convolutional layers and full connection layers are constructed as follows.
The first convolutional layer contains 32 filters.
Each filter's size is $4\times4$ and it moves $2\times2$ stride every time through the full depth of the input data.
The second convolutional layer has 64 filters.
Each filter's size is $2\times2$ and it moves $2\times2$ stride every time.
The size of the output after two convolutional layers is $15\times15\times64$.
The third convolutional layer has 128 filters with the size of $2\times2$ and the stride's size is $1\times1$.
The third convolutional layer's output is a $15\times15\times128$ tensor.
A fully-connected layer transfers the tensor into a $128\times1$ matrix.
After the fully-connected layer, the data are split into two parts with the same size $64\times1$.
The first part is then used to calculate the value and the second part is for the advantage.
The advantage of an action means how well it can achieve by taking an action over all the other actions.
Because the number of possible actions in our system is 9 as shown in Fig. \ref{fig:mdp2}, the size of the advantage is $9\times1$.
They are combined again to get the $Q$ value, which is the architecture of the dueling Deep Q Network (DQN).


With the $Q$ value corresponding to every action, we need highly penalize illegal actions, which may cause accidents or reach the max/min signal duration.
The output combines the $Q$ value and tentative actions to force the traffic light to take a legal action. 
Finally we get the $Q$ values of every action in the output with penalized values.
The parameters in the CNN is denoted by $\theta$.
$Q(s, a)$ now becomes $Q(s, a; \theta)$, which is estimated under the CNN $\theta$.
The details in the architecture are presented in the next subsections.

\subsection{Dueling DQN}
As mentioned before, our network contains a dueling DQN\cite{WanSch15}. 
In the network, the $Q$ value is estimated by the value at the current state and each action's advantage compared to other actions. 
The value of a state $V(s;\theta)$ denotes the overall expected rewards by taking probabilistic actions in the future steps.
The advantage corresponds to every action, which is defined as $A(s,a;\theta)$.
The $Q$ value is 
the sum of the value $V$ and the advantage function $A$, which is calculated by the following equation,
\begin{equation}
    \begin{split}
        Q(s,a;\theta) = &V(s;\theta)+ \\&\left(A(s,a;\theta)-\frac{1}{|A|}\sum_{a'}A(s,a';\theta)\right).
    \end{split}
\end{equation}
$A(s,a;\theta)$ 
shows how important an action is to the value function among all actions.
If the $A$ value of an action is positive, it means the action shows a better performance in numerical rewards compared to the average performance of all possible actions;
otherwise, if the value of an action is negative, it means the action's potential reward is less than the average.
It has been shown that the subtraction from the mean of all advantage values can improve the stability of optimization compared to using the advantage value directly.
The dueling architecture is shown to effectively improve the performance in reinforcement learning.

\subsection{Target Network}
To update the parameters in the neural network, a target value is defined to help guide the update process.
Let $Q_{target}(s,a)$ denote the target Q value at the state $s$ when taking action $a$.
The neural network is updated by the Mean Square Error (MSE) in the following equation,
\begin{equation}
\label{equ:mse}
J = \sum_{s} P(s)[Q_{target}(s, a)-Q(s,a;\theta)]^2,
\end{equation}
where $P(s)$ denotes the probability of state $s$ in the training mini-batch.
The MSE can be considered as a loss function to guide the updating process of the primary network.
To provide stable update in each iteration, a separate target network $\theta^-$, the same architecture as the primary neural network but different parameters, is usually employed to generate the target value.
The calculation of the target $Q$ value is presented in the double DQN part.

The parameters $\theta$ in the primary neural network are updated by back propagation with (\ref{equ:mse}).
$\theta^-$ is updated based on the $\theta$ in the following equation,
\begin{equation}
\label{equ:u2}
\theta^- = \alpha \theta^-+(1-\alpha)\theta.
\end{equation}
$\alpha$ is the update rate, which presents how much the newest parameters affect the components in the target network. 
A target network can help mitigate the overoptimistic value estimation problem.

\begin{figure*}[!t]
\centerline{ \psfig{figure=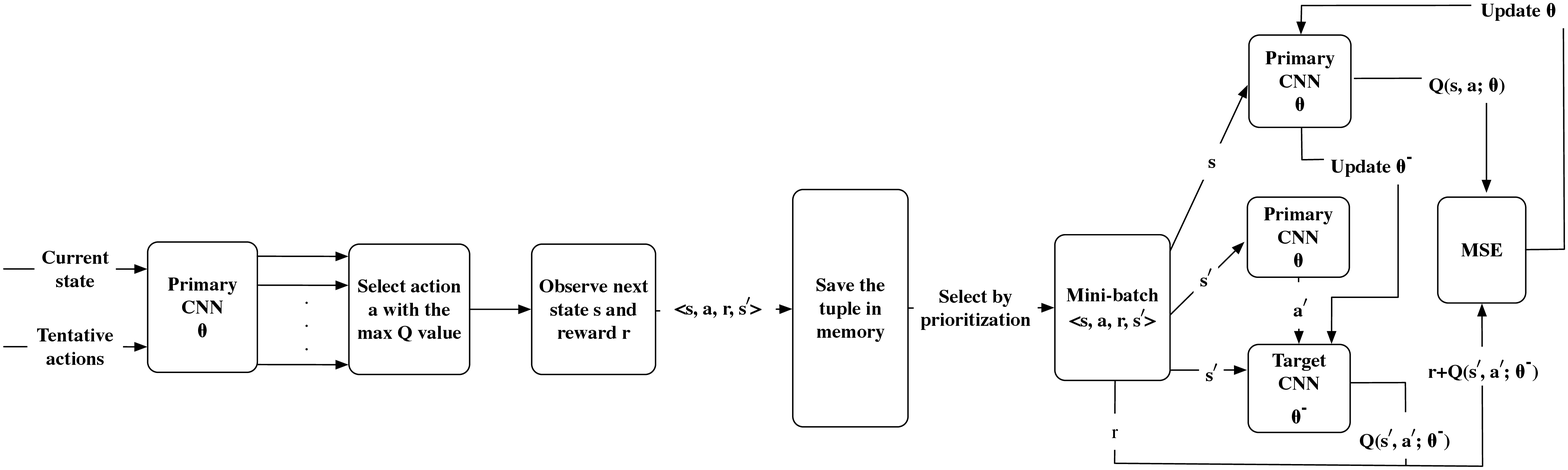,width=6.5in}}
\caption{The architecture of the reinforcement learning model in our system}
\label{fig:arch}
\end{figure*}

\subsection{Double DQN}
The target Q value is generated by the double Q-learning algorithm \cite{VanGue16}.
In the double DQN, the target network is to generate the target $Q$ value and the action is generated from the primary network.
The target $Q$ value can be expressed in the following equation,
\begin{equation}
\label{equ:qt}
    Q_{target}(s, a) = r+\gamma Q(s', \mathop{\arg \max}\limits_{a'}(Q(s', a'; \theta)), \theta^-).
\end{equation}
It is shown that the double DQN effectively mitigates the overestimations and improves the performance \cite{VanGue16}.

In addition, we also employ the $\epsilon$-greedy algorithm to balance the exploration and exploitation in choosing actions. 
With the increasing steps of training process, the value of $\epsilon$ decreases gradually.
We set a starting and ending values of $\epsilon$ and the number of steps to reach the ending value.
The value of $\epsilon$ linearly decreases to the ending value.
When $\epsilon$ reaches the ending value, it keeps the value in the following procedure.



\subsection{Prioritized Experience Replay}
During the updating process, the gradients are updated through the experience replay strategy.
A prioritized experience replay strategy chooses samples from the memory based on priorities, which can lead to faster learning and to better final policy\cite{SchQua15}.
The key idea is to increase the replay probability of the samples that have a high temporal difference error.
There are two possible methods estimating the probability of an experience in a replay, proportional and rank-based.
Rank-based prioritized experience replay can provide a more stable performance since it is not affected by some extreme large errors.
In this system, we take the rank-based method to calculate the priority of an experience sample.
The temporal difference error $\delta$ of an experience sample $i$ is defined in the following equation, 
\begin{equation} 
    \delta_i = |Q(s,a;\theta)_i-Q_{target}(s, a)_i|.
\end{equation}
The experiences are ranked by the errors and then the priority $p_i$ of  experience $i$ is the reciprocal of its rank.
Finally, the probability of sampling the experience $i$ is calculated in the following equation,
\begin{equation} 
    P_i = \frac{p_i^\tau}{\sum_k p^\tau_k}.
\end{equation}
$\tau$ presents how much prioritization is used.
When $\tau$ is 0, it is random sampling.

\subsection{Optimization}
In this paper, we optimize the neural networks by the ADAptive Moment estimation (Adam) \cite{KinBa14}.
The Adam is evaluated and compared with other back propagation optimization algorithms in \cite{Rud16}, which concludes that 
the Adam attains satisfactory overall performance with a fast convergence and adaptive learning rate.
The Adam optimization method adaptively updates the learning rate considering both first-order and second-order moments using the stochastic gradient descent procedure.
Specifically, let $\boldsymbol{\theta}$ denote the parameters in the CNN and $J(\boldsymbol{\theta})$ denote the loss function.
Adam first calculates the gradients of the parameters, 
\begin{equation} 
    \textbf{g} = \nabla_\theta J(\boldsymbol{\theta}). 
\end{equation}
It then respectively updates the first-order and second-order biased moments, $\textbf{s}$ and $\textbf{r}$, by the exponential moving average, 
\begin{equation}
    \begin{split}
        \textbf{s}  = \rho_s \textbf{s}+(1-\rho_s)\textbf{g},\\
        \textbf{r}  = \rho_r \textbf{r}+(1-\rho_r)\textbf{g},
    \end{split}
\end{equation}
where $\rho_s$ and $\rho_r$ are the exponential decay rates for the first-order and second-order moments, respectively.
The first-order and second-order biased moments are corrected using the time step $t$ through the following equations, 
\begin{equation}
    \begin{split}
        \hat{\textbf{s}}  = \frac{\textbf{s}}{1-\rho_s^t},\\
        \hat{\textbf{r}}  = \frac{\textbf{r}}{1-\rho_r^t}.
    \end{split}
\end{equation}
Finally the parameters are updated as follows,
\begin{equation}
    \begin{split}
        \boldsymbol{\theta}  = & \boldsymbol{\theta}+\Delta\boldsymbol{\theta} \\
        = & \boldsymbol{\theta}+\left(-\epsilon_r \frac{\hat{\textbf{s}}}{\sqrt{\hat{\textbf{r}}+\delta}}\right),
    \end{split}
\end{equation}
where $\epsilon_r$ is the initial learning rate and $\delta$ is a small positive constant to attain numerical stability.

\subsection{Overall Architecture}

In summary, the whole process in our model is shown in Fig. \ref{fig:arch}.
The current state and the tentative actions are fed to the primary convolutional neural network to choose the most rewarding action.
The current state and action along with the next state and received reward are stored into the memory as a four-tuple $\left < s, a, r, s'\right>$.
The data in the memory are selected by the prioritized experience replay to generate mini-batches and they are used to update the primary neural network's parameters.
The target network $\theta^-$ is a separate neural network to increase stability during the learning.
We use the double DQN \cite{VanGue16} and dueling DQN \cite{WanSch15} to reduce the possible overestimation and improve performance.
Through this way, the approximating function can be trained and the $Q$ value at every state to every action can be calculated.
The optimal policy can then be obtained by choosing the action with the max $Q$ value.

\begin{algorithm}
    \caption{Dueling Double Deep Q Network with Prioritized Experience Replay Algorithm on a Traffic Light}
\label{alg:trn}
{\fontsize{9}{10}\selectfont
\begin{algorithmic}

\STATE{Input:} replay memory size $M$, minibatch size $B$, greedy $\epsilon$, pre-train steps $tp$, target network update rate $\alpha$, discount factor $\gamma$.
\STATE \underline{Notations:}
    \STATE $\theta$: the parameters in the primary neural network.
    \STATE $\theta^-$: the parameters in the target neural network.
    \STATE $m$: the replay memory.
    \STATE $i$: step number.
\vspace{0.3cm}

\STATE Initialize parameters $\theta$, $\theta^-$ with random values.
\STATE Initialize $m$ to be empty and $i$ to be zero. 
\STATE Initialize $s$ with the starting scenario at the intersection.
\WHILE{there exists a state $s$} 
    \STATE Choose an action $a$ according to the $\epsilon$ greedy.
    \STATE Take action $a$ and observe reward $r$ and new state $s'$.
    \IF{the size of memory $m > M$}
    \STATE Remove the oldest experiences in the memory.
    \ENDIF
    \STATE Add the four-tuple $\left \langle s, a, r, s' \right \rangle$ into $M$.
    \STATE Assign $s'$ to $s$: $s$ $\gets$ $s'$.
    \STATE $i\gets i+1$.
    \IF{$|M| > B$ and $i > tp$}
    \STATE Select $B$ samples from $m$ based on the sampling priorities.
    \STATE Calculate the loss $J$: 
    \STATE{ \quad$\begin{aligned} J = &\sum_{s} \frac{1}{B}[r+\gamma Q(s', \mathop{\arg \max}\limits_{a'}(Q(s', a'; \theta)), \theta^-)-\\ &Q(s,a;\theta)]^2. \end{aligned}$}
    \STATE Update $\theta$ with $\nabla J$ using Adam back propagation.
    \STATE Update $\theta^-$ with $\theta$: 
    \STATE \quad $\theta^- = \alpha \theta^-+(1-\alpha)\theta.$
    \STATE Update every experience's sampling priority based on $\delta$.
    \STATE Update the value of $\epsilon$.
    \ENDIF
\ENDWHILE
\end{algorithmic}}
\end{algorithm}

The pseudocode of our 3DQN with prioritized experience replay is shown in Algorithm \ref{alg:trn}.
Its goal is to train a mature adaptive traffic light, which can change its phases' duration based on different traffic scenarios.
The agent first chooses actions randomly till the number of steps is over the pre-train steps and the memory has enough samples for at least one mini-batch.
Before the training, every samples' priorities are the same.
Thus, they are randomly selected into a mini-batch to train.
After training once, the samples' priorities change and they are selected by different probabilities.
The parameters in the neural network is updated by the Adam back propagation\cite{Rud16}.
The agent chooses actions based on the $\epsilon$ and the action that has the max $Q$ value.
The agent finally learns to get a high reward by reacting on different traffic scenarios.


\section{evaluation}
\label{sec:eva}

In this section, we present the simulation environment.
Our proposed model is then evaluated via simulation, and the simulation results are presented to show the effectiveness of our model.

\subsection{Evaluation Methodology and Parameters}

Our main objective in conducting the simulation is as follows,
\begin{itemize}
\item Maximizing the defined reward, which is to reduce the cumulative delay of all vehicles.
\item Reducing the average waiting time of vehicles in the traffic road scenario.
\end{itemize}
To specifically, the first objective is the goal of a reinforcement learning model.
We measure the cumulative reward in every episode within one hour period.
The second objective is an important metric in measuring the performance of a traffic management system, which directly affects the drivers' feelings.
For the both objectives, we compare the performance of the proposed model with pre-scheduled traffic signals.
At intersections with traditional traffic lights, the signals are pre-scheduled by the operator and they do not change any more.


\begin{figure}
\centerline{ \psfig{figure=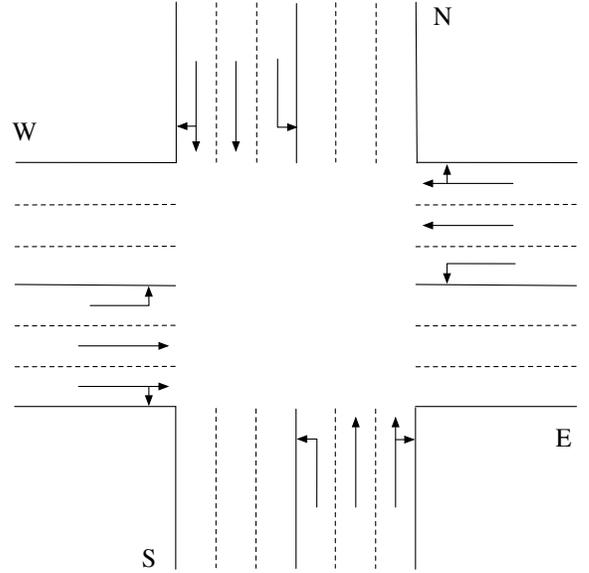,width=3.0in}}
    \caption{The intersection scenario tested in our evaluation.}
\label{fig:ints}
\end{figure}
The evaluation is conducted in SUMO \cite{KraErd12}, which provides real-time traffic simulation in a micro way.
We use the Python APIs provided by SUMO to extract the traffic light controlled intersection's information and to send orders to change the traffic light's timing.
The intersection is composed of four perpendicular roads, which is shown in Fig. \ref{fig:ints}.
Every road has three lanes.
The right-most lane allows the right-turn and through traffic, the middle one is the through only lane, and the left inner lane allows the left-turn vehicles only.
The whole intersection scenario is a $300m\times 300m$ area.
The lane length is $150$ meters.
The vehicle length is 5 meters and the minimal gap between two vehicles is 2 meters.
We set the grid length $c$ 5 meters, thus the total number of grids is $60\times 60$.
The vehicles arrive in the scenario following a random process.
The average vehicle arrival rate of every lane is the same, 1/10 per second. 
There are two through lanes, so the flow rate of all through traffic (west-to-east, east-to-west, north-to-south, south-to-north) is 2/10 per second, and the turning traffic (east-to-south, west-to-north, south-to-west, north-to-east) is 1/10 per second. 
SUMO provides the Krauss Following Model \cite{Kra97}, which guarantees the safe driving on the road.
For vehicles, the max speed is 13.9 $m/s$, which is equal to 50 $km/h$. 
The max accelerating acceleration is 1.0 $m/s^2$ and the decelerating acceleration is 4.5 $m/s^2$.
The duration of yellow signals $T_{yellow}$ is set $4$ seconds.

The model is trained in iterations.
One iteration is an episode with traffic in an hour.
The reward is accumulated in an episode.
The goal in our network is to maximize the reward in the one-hour episode by modifying the traffic signals' time duration.
The simulation results are the average values of the nearest 100 iterations.
The development environment is built on the top of Tensorflow \cite{AbaAga16}.
The parameters in the network are shown in Table \ref{tbl:prm}.
The performance in our system is first compared with the traffic lights with fix-time signals.
We fix the traffic signals' time duration as 30 seconds and 40 seconds.
The model is then compared to other deep reinforcement learning architectures with different parameters.

\begin{table} 
\centering
\caption{\textbf{Parameters in the reinforcement learning network}}
\label{tbl:prm}
\begin{tabular}{lc}
    \Xhline{3\arrayrulewidth}
 Parameter & Value \\
\hline
 Replay memory size $M$ &  20000 \\
 Minibatch size $B$ &  64 \\
 Starting $\epsilon$  &  1 \\
 Ending $\epsilon$  &  0.01 \\
 Steps from starting $\epsilon$ to ending $\epsilon$ &  10000 \\
 Pre-training steps $tp$  &  2000 \\
 Target network update rate $\alpha$ &  0.001 \\
 Discount factor $\gamma$ & 0.99 \\
 Learning rate $\epsilon_r$&  0.0001 \\
 Leaky ReLU $\beta$&  0.01 \\
    \Xhline{3\arrayrulewidth}

\end{tabular}
\end{table}

\subsection{Experimental Results}

\subsubsection{Cumulative reward}
\begin{figure} 
\centerline{ \psfig{figure=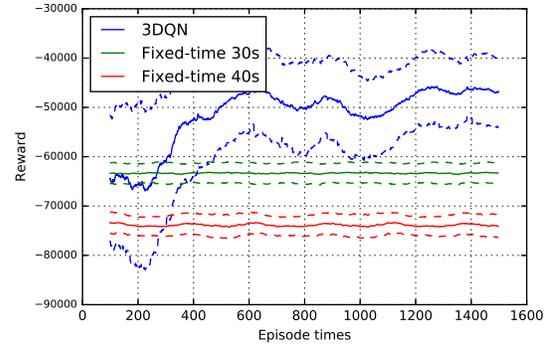,width=3.0in}}
    \caption{The cumulative reward during all the training episodes.}
\label{fig:rwd}
\end{figure}

The accumulated reward in every episode is first evaluated with the same traffic flow rate from all lanes.
The simulation results are shown in Fig. \ref{fig:rwd}.
The blue real line shows the results in our model and the green and red real lines are the results from fixed-time traffic lights.
The dotted lines are the corresponding confidence intervals of the corresponding color's real lines.
From this figure, we can see that our 3DQN outperforms the other two strategies with fixed-time traffic lights.
Specifically, the cumulative reward in one iteration is greater than -50000 (note that the reward is negative since the vehicles' delay is positive) while that in the other two strategies is less than -6000.
The fixed-time traffic signals always obtains a low reward even though more iterations are generated while
our model can learn to achieve a higher reward with more iterations.
This is because the fixed-time traffic signals do not change the signals' time under different traffic scenario. 
In the 3DQN, the signals' time changes to achieve the best expected rewards, which balances the current traffic scenario and the potential future traffic.
When the training process iterates over 1000 times in our protocol, the cumulative rewards become more stable than previous iterations.
It means the protocol has learnt how to handle different traffic scenarios to get the most rewards after 1000 iterations.

\subsubsection{Average waiting time}
\begin{figure}
\centerline{ \psfig{figure=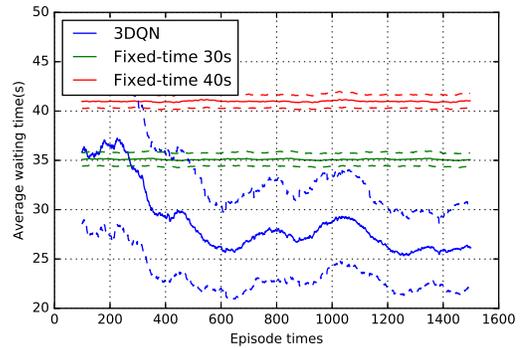,width=3.0in}}
    \caption{The average waiting time during all the training episodes.}
\label{fig:awt}
\end{figure}

We test the average waiting time of vehicles in every episode, which is shown in Fig. \ref{fig:awt}.
In this scenario, the traffic rates from all lanes are also the same.
In this figure, the blue real line shows the results in our model, and the green and red real lines are the results from fixed-time traffic lights.
Also the dotted lines are corresponding variances of the same color's dot lines.
From this figure, we can see that our 3DQN outperforms the other two strategies with fixed-time traffic lights.
Specifically, the average waiting time in the fixed-time signals is always over 35 seconds.
Our model can learn to reduce the waiting time to about 26 seconds after iterating 1200 times from over 35 seconds, which is at least 25.7\% less than the other two strategies.
It shows that our model can greatly improve the performance in vehicles' average waiting time at intersections. 

\subsubsection{Comparison with different parameters and algorithms}
\begin{figure}
\centerline{ \psfig{figure=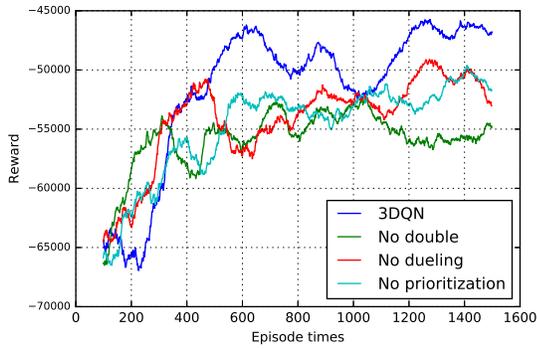,width=3.0in}}
    \caption{The cumulative reward during all the training episodes in different network architecture.}
\label{fig:rwd-p}
\end{figure}

In this part, we evaluate our model by comparing to others with different parameters.
In our model, we used a series of techniques to improve the performance of deep Q networks.
For comparison, we remove one of these techniques each time to see how the removed technique affects the performance.
The techniques include double network, dueling network and prioritized experience replay.
We evaluate them by comparing the performance with the employed model.
The reward changes in all methods are shown in Fig. \ref{fig:rwd-p}.
The blue real line presents our model, and the green line is the model without double network.
The red line is the model without dueling network and the cyan line is the model without prioritized experience replay.
We can see that our model can learn fastest among the four models.
It means our model reaches the best policy faster than others.
Specifically, even there is some fluctuation in the first 400 iterations, our model still outperforms the other three after 500 iterations. 
Our model can achieve greater than -47000 rewards while the others have less than -50000 rewards.

\subsubsection{Average waiting time under rush hours}
\begin{figure}
\centerline{ \psfig{figure=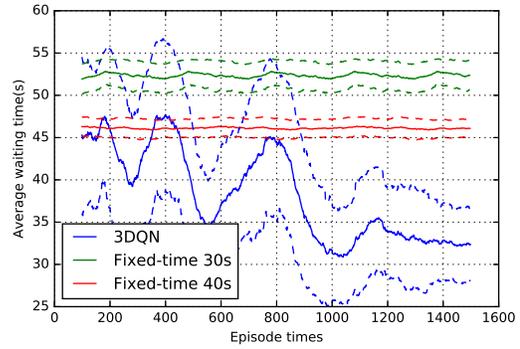,width=3.0in}}
    \caption{The average waiting time in all the training episodes during the rush hours with unbalanced traffic from all lanes.}
\label{fig:awt-u}
\end{figure}

In this part, we evaluate our model by comparing the performance under the rush hours.
The rush hour means the traffic flows from all lanes are not the same, which is usually seen in the real world.
During the rush hours, the traffic flow rate from one direction doubles, and the traffic flow rates in the other lanes keep the same as normal hours.
Specifically, in our experiments, the arrival rate of vehicles on the lanes from the west to east becomes $2/10$ each second and the arrival rates of vehicles on the other lanes are still $1/10$ each second.
The experimental result is shown in Fig. \ref{fig:awt-u}.
In this figure, the blue real line shows the results in our model and the green and red real lines are the results from fixed-time traffic lights.
The dotted lines are the corresponding variances of the corresponding color's real lines.
From the figure, we can see that the best policy becomes harder to be learnt than the previous scenario. 
This is because the traffic scenario becomes more complex, which leads to more uncertain factors.
But after trial and error, our model can still learn a good policy to reduce the average waiting time.
Specifically, the average waiting time in 3DQN is about 33 seconds after 1000 episodes while the average waiting time in the other two methods is over 45 seconds and over 50 seconds.
Our model reduces about 26.7\% of the average waiting.

\section{Conclusion}
\label{sec:con}
In this paper, we propose to solve the traffic light control problem using the deep reinforcement learning model. 
The traffic information is gathered from vehicular networks. 
The states are two-dimension values with the vehicles' position and speed information.
The actions are modeled as a Markov decision process and the rewards are the cumulative waiting time difference between two cycles.
To handle the complex traffic scenario in our problem, we propose a double dueling deep Q network (3DQN) with prioritized experience replay.
The model can learn a good policy under both the rush hours and normal traffic flow rates.
It can reduce over 20\% of the average waiting timing from the starting training.
The proposed model also outperforms others in learning speed, which is shown in extensive simulation in SUMO and TensorFlow.

\bibliographystyle{IEEEtran}
\bibliography{ref}
\end{document}